\documentclass[10pt,twocolumn,letterpaper]{article}

\usepackage[pagenumbers]{cvpr}      %

\usepackage{cuted} %

\usepackage{currfile} %

\usepackage{caption} %

\usepackage{pifont}   %
\usepackage{multirow}
\newcommand{\cmark}{\ding{51}}
\newcommand{\xmark}{\ding{55}}
\newcommand{\ours}{PCW\xspace}
\newcommand{\oursfull}{Predictive Code World\xspace}
\newcommand{\flash}{Qwen3.8-Flash\xspace}
\newcommand{\qmax}{Qwen3.8-Max\xspace}
\newcommand{\pass}{\texttt{pass}\xspace}
\newcommand{\fail}{\texttt{fail}\xspace}
\newcommand{\unknown}{\texttt{unknown}\xspace}

\makeatletter
\@ifpackageloaded{lineno}{%
  \ifdefined\IfFormatAtLeastTF
    \IfFormatAtLeastTF{2025-06-01}{%
      \@ifpackagelater{lineno}{2026/02/11}{\RemoveFromHook{build/column/before}[lineno]}{}%
      \AddToHook{build/column/after}{%
        \setbox\@outputbox\vbox{%
          \boxmaxdepth\@maxdepth
          \protected@write\@auxout{}{\string\def\string\@LN@column{\if@firstcolumn1\else2\fi}}%
          \box\@outputbox}}%
    }{}%
  \fi
}{}
\makeatother

\usepackage{microtype}

\definecolor{cvprblue}{rgb}{0.21,0.49,0.74}
\usepackage[pagebackref,breaklinks,colorlinks,allcolors=cvprblue]{hyperref}

\makeatletter
\patchcmd{\@maketitle}{\vskip .375in}{\vskip .18in}{}{}
\patchcmd{\@maketitle}{\vspace*{24pt}}{\vspace*{7pt}}{}{}
\patchcmd{\@maketitle}{\vspace*{12pt}\end{center}}{\vspace*{4pt}\end{center}}{}{}
\makeatother

\title{Predictive Code Worlds: Write the World First,\\
Then Predict It, for Robot Code Agents}
\author{%
  Ge Wang\textsuperscript{1}\quad Zhixin Mai\textsuperscript{1}\quad
  Zheng SUN\textsuperscript{2}\quad Mingkai Jia\textsuperscript{3}\quad
  Ziyue Zhu\textsuperscript{4}\quad Zhen Li\textsuperscript{1}\\[3pt]
  {\small \textsuperscript{1}The Chinese University of Hong Kong, Shenzhen
  \quad \textsuperscript{2}The Chinese University of Hong Kong}\\
  {\small \textsuperscript{3}The Hong Kong University of Science and Technology
  \quad \textsuperscript{4}Nankai University}%
}
\hypersetup{
  pdftitle={Predictive Code Worlds: Write the World First, Then Predict It, for Robot Code Agents},
  pdfauthor={Ge Wang, Zhixin Mai, Zheng SUN, Mingkai Jia, Ziyue Zhu, Zhen Li}
}
\date{November 2026}

\begin{document}
\maketitle
\begin{strip}
\centering
\includegraphics[width=\linewidth]{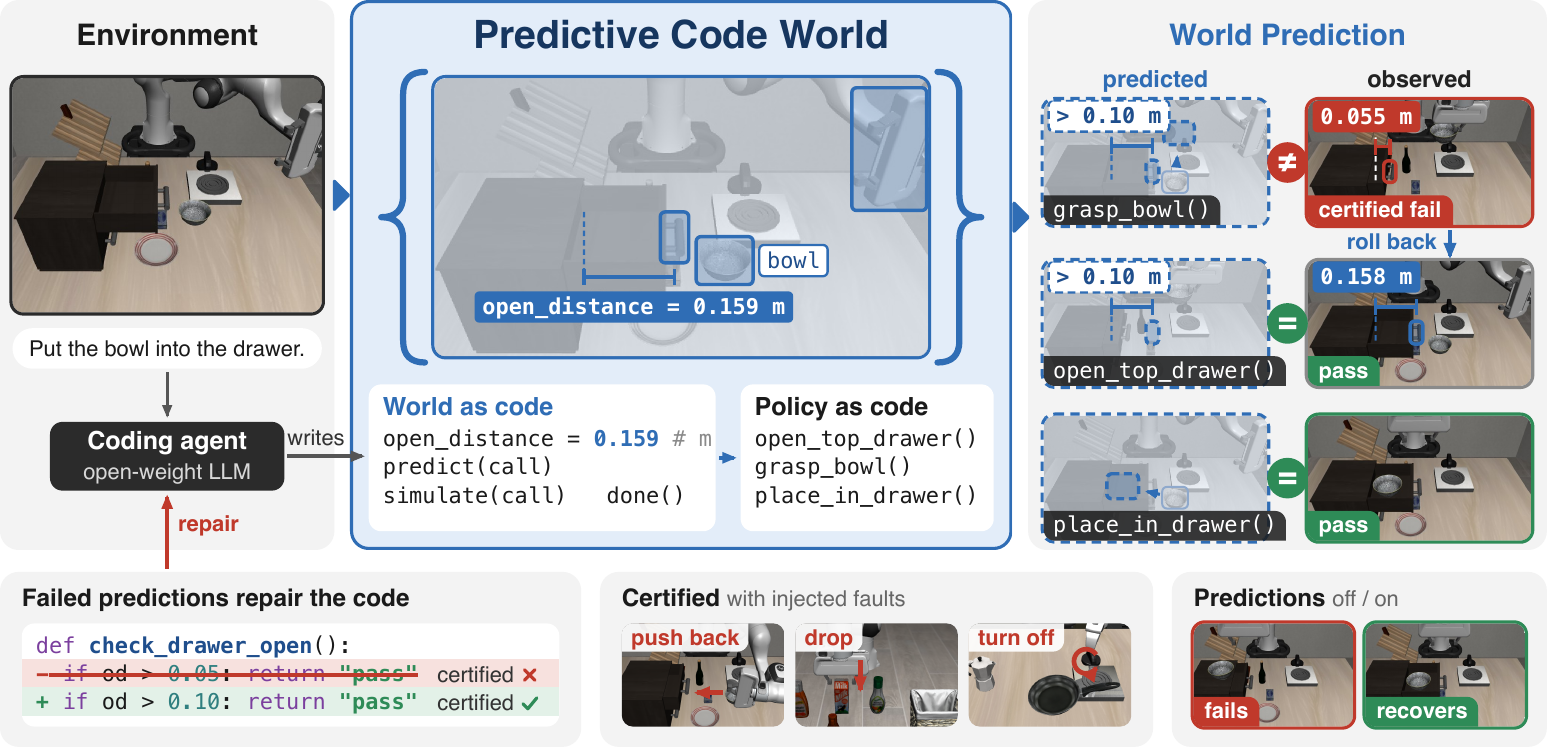}
\captionsetup{hypcap=false}
\captionof{figure}{%
\textbf{\oursfull (\ours).}
\emph{Left:} given one scene and one instruction, a coding agent writes two programs.
\emph{Center:} it first writes the scene as code, a \emph{code world} of named things and measured facts with their units, and only then its policy on top of it.
\emph{Right:} for each subgoal the code world predicts the world that should follow (dashed); when a fresh observation disagrees and the prediction is certified, the program rolls back, here to re-open a drawer that was silently pushed back.
\emph{Bottom:} failed predictions make the agent repair its code during development; predictions are certified with injected faults before they may act (three tasks shown); without them the same episode fails.
Blue pictures are drawings of the code world's facts, which are code, not images.}
\captionsetup{hypcap=true}
\label{fig:\currfilebase}%
\end{strip}

\begin{abstract}
Code-as-policy agents write robot programs on top of perception and motion APIs, and how well they do depends heavily on the language model behind them. Masks, depth and bounding boxes are not a model of the task's world, so the model has to work that world out by itself.
We present \emph{\oursfull} (\ours), in which the agent writes the world down before the policy and then makes that world predictive.
The agent first writes a \emph{code world}, an executable model of its task that keeps measured facts, predicts what each action should change and judges progress toward the goal, and writes its skill on it.
Once the skill works, the code world also predicts the observable outcome of each subgoal.
These \emph{world predictions} come from the same agent as the skill they check and are not trusted by default: faults injected during development label their verdicts, and only predictions certified on these labels may roll the program back.
With a small open-weight model, \flash, and self-improvement, \ours reaches 89.4\% success on 80 LIBERO-Pro tasks, surpassing ASPIRE with the same model, a larger open-weight model, and even the closed-source Claude Opus 4.6; its gains carry over to Robosuite and BEHAVIOR-1K.
Under disturbances, world predictions catch failures that would otherwise go unnoticed and recover from them.
\end{abstract}

\section{Introduction}
\label{sec:intro}

Robot behavior can be written as a program: code-as-policy agents compose perception and motion APIs into executable policies~\cite{codeaspolicies,progprompt}, and because the behavior is code, recent agents inspect execution traces and repair their own programs~\cite{aspire}.
With the strongest frontier model, such an agent almost always succeeds on LIBERO-Pro~\cite{aspire,liberopro}.
However, its success follows the strength of its model: with weaker models the same agent succeeds far less often, and a small open-weight model, without self-improvement, succeeds in fewer than half of the episodes.

We trace this gap to what the program is written on.
Consider putting a bowl into a drawer (\cref{fig:teaser}).
Perception APIs return masks, depth and bounding boxes, not a model of the task's world: which mask is the drawer handle, how far the drawer has been pulled, whether the bowl moved with the hand.
A program written directly on these outputs has to rediscover such facts wherever it needs them, and when the drawer is pushed back before the bowl arrives, nothing in the program says what the world should have looked like.
A strong model makes up for this by itself; a weaker one does not.
We therefore ask: can the agent write down the world its program acts in, so that a small model can write a good skill on it, and so that the program notices when the world goes wrong?

We present \emph{\oursfull} (\ours), which answers both halves of this question (\cref{fig:teaser}).
World models written as code are by now an active line of work~\cite{worldcoder,gifmcts,poeworld,codeasworlds}; \ours goes further with its two parts.
\emph{(1) A code world to write the skill on.}
For each task, the agent begins with a \emph{code world}, an executable model of this one scene and instruction: it keeps the task's facts with their units and evidence, predicts what each action should change, rehearses candidate programs offline and judges how far away the goal is.
The agent then writes its \emph{skill}, the policy program, on this model instead of on raw perception.
\emph{(2) World predictions that are certified before they act.}
When the skill works, the agent adds predictions about outcomes: for each subgoal, what a fresh observation should show after success and after the most likely failure.
These \emph{world predictions} are written by the same agent as the skill they check, so they share its blind spots: a threshold the agent misjudged in the skill is misjudged in the prediction too.
And a wrong verdict is not harmless, since acting on it can undo a skill that was working.
We therefore do not trust them by default.
We inject faults during development, a dropped object or a part moved back, to label their verdicts, and only verdicts that prove precise on these labels are certified and may act: at run time, a certified failure rolls the program back to the earliest subgoal that no longer holds.

We evaluate \ours on LIBERO-Pro, Robosuite and BEHAVIOR-1K.
On 80 LIBERO-Pro tasks, \flash with \ours reaches 89.4\%, whereas ASPIRE reaches 63.8\% with the same model and 65.7\% with Claude Opus 4.6.
Most of the gain comes from the code world: without self-improvement, writing on a code world instead of on raw perception takes the same model from 49.4\% to 78.2\%.
World predictions add to it on all three benchmarks, and most under disturbances: they catch failures that the program would otherwise never notice, and nearly triple success when a subgoal is undone after it was confirmed.
These recoveries depend on the injected faults, without which almost no \fail verdict could have been certified.

\noindent Our contributions are:
\begin{itemize}
\item \textbf{Code worlds} as the representation on which a robot code agent writes its skill.
\item \textbf{World predictions} at the level of subgoals, certified on faults injected during development before they may act.
\item \textbf{Evidence} on three benchmarks that a small open-weight model can then outperform agents built on larger and stronger models and recover from disturbances.
\end{itemize}

\section{Related Work}
\label{sec:related}

\paragraph{Code as policies and robot coding agents.}
End-to-end vision-language-action policies map images and instructions directly to actions~\cite{openvla,pi0}, but they degrade under the perturbations of LIBERO-Pro~\cite{liberopro}.
Code-as-policy systems take another route: a language model writes a robot program that composes perception and control primitives~\cite{codeaspolicies,progprompt,voxposer}, or states the task's constraints as code~\cite{rekep}.
Coding agents now debug such programs from execution feedback and improve them across attempts~\cite{selfdebug,reflexion,voyager,eureka}.
For robots, CaP-X~\cite{capx} benchmarks them on manipulation, RAPID~\cite{rapid} refines programs generated from a human demonstration, Playful RATs~\cite{rats} learn skills from self-proposed goals, and ASPIRE~\cite{aspire} repairs programs from execution traces and accumulates a skill library, and almost always succeeds on LIBERO-Pro with frontier models.
\ours keeps this way of developing programs and changes what the program is written on, a code world, whose world predictions act only once they are certified.

\paragraph{World models.}
Learned world models predict future observations or latent states and support planning and policy learning~\cite{worldmodels,dreamer,dreamerv3,unisim,genie,cosmos,vjepa2,dinowm}.
A second line writes world models as code: programs that represent state and transitions are synthesized and refined from interaction~\cite{worldcoder,gifmcts,poeworld,codeasworlds}.
HexaAnything~\cite{hexaanything} pairs a code world with a code policy and uses provenance records and predicate verification for recovery and evolution; like it, \ours writes both its world and its policy as code.
Other recent agents keep executable models that predict before acting and revise them from experiments, counterexamples and replay~\cite{empiric,memento3,kepler}.
Our code world belongs to the second line, and goes beyond it in two ways.
It is written by the agent for one task as the representation its skill is written on: it builds its state from perception APIs, with units, frames and evidence for every fact, and its action-level predictions and rehearsal model serve the development of that skill.
And its world predictions, which state the observable outcome of each subgoal's success and most likely failure, are certified on injected faults before they may act.

\paragraph{Failure detection and recovery.}
Manipulation systems detect failures with vision-language success detectors~\cite{successdetectors,aha}, explain and correct them from summarized experience~\cite{reflect}, detect plan--execution misalignment~\cite{doremi}, or close the loop with language feedback~\cite{innermonologue}.
Code-as-Monitor~\cite{codeasmonitor} has a vision-language model formulate spatio-temporal constraints and evaluates them as code for reactive and proactive failure detection, and VASO~\cite{vaso} turns formal counterexamples into textual gradients that refine reusable skill contracts.
Our world predictions also check outcomes at run time, as code and without querying a language model. What differs is how they come to act: each verdict is labeled with injected faults during development and acts only once it is certified, and completed subgoals are checked again so that the program can roll back to the earliest one undone.

\begin{figure*}[t]
\centering
\includegraphics[width=\linewidth]{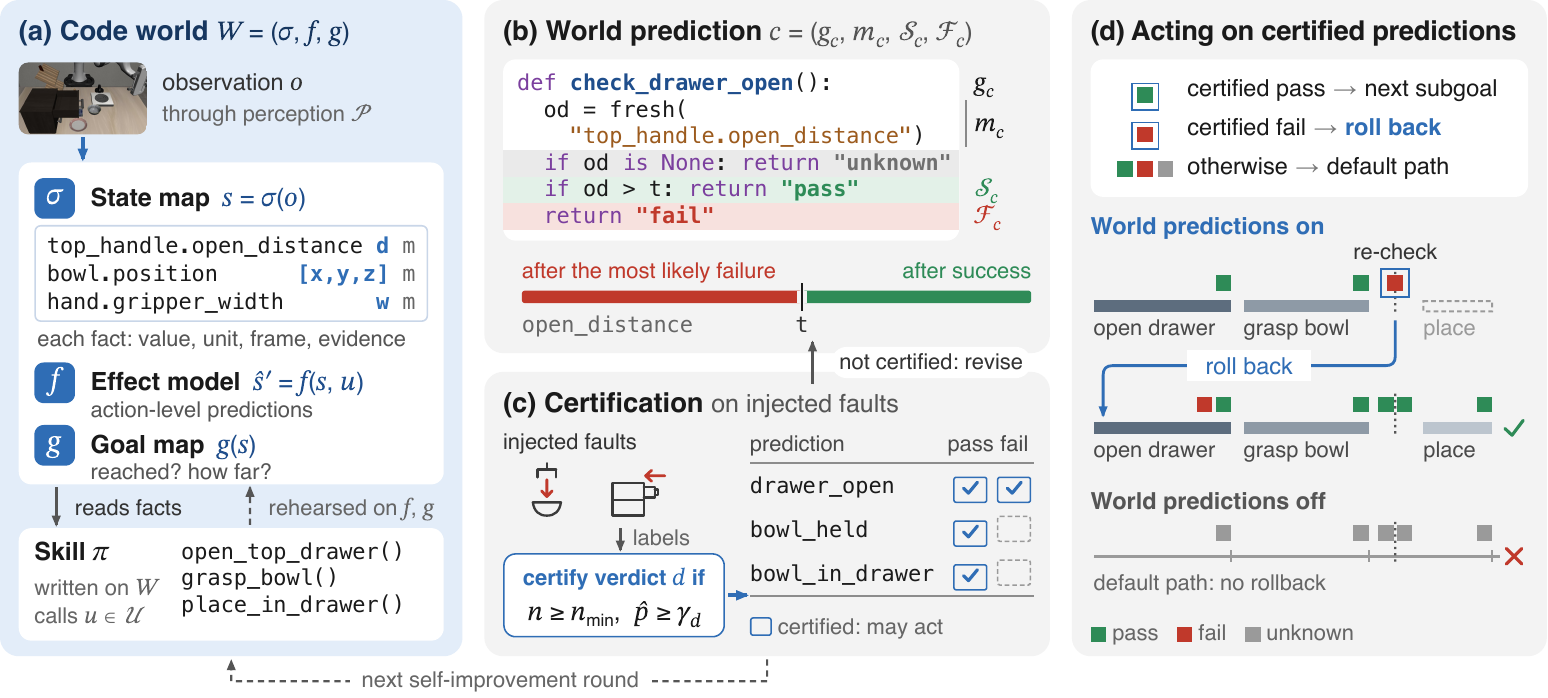}
\caption{\textbf{Overview of \ours.}
\textbf{(a)}~The agent first writes a code world $W=(\sigma,f,g)$: a state map that turns an observation into named facts, an effect model that predicts what a motion call should change, and a goal map. The skill~$\pi$ reads its facts and is rehearsed on it.
\textbf{(b)}~A world prediction measures one quantity for a subgoal and names the range expected after success and the range expected after the most likely failure (\cref{eq:settle}); here a threshold $t$ separates them.
\textbf{(c)}~Faults injected during development label its verdicts, and each of its two verdicts, \pass and \fail, is certified only if it is precise enough on these labels (\cref{eq:certify}); a verdict that is not certified sends the agent back to revise.
\textbf{(d)}~At run time only certified verdicts act: when a re-check of the completed subgoals finds a certified \fail, the program rolls back to the earliest subgoal that no longer holds; otherwise, and with world predictions off, it stays on its default path.
The example is the bowl-into-drawer program, simplified; the figure is schematic and shows no logged values.}
\label{fig:overview}
\end{figure*}

\paragraph{Calibration, selective prediction and fault injection.}
Acting on a predictor only where it is reliable is the idea behind calibration~\cite{calibration}, selective prediction~\cite{selective} and conformal prediction~\cite{conformal}, which KnowNo~\cite{knowno} uses to decide when a language-model planner should ask for help.
Language models can partly estimate their own correctness~\cite{lmknow}, yet they struggle to correct their own reasoning without external feedback~\cite{selfcorrect}; a self-written world model should therefore be checked against outcomes, not against its author's confidence.
Estimating how reliable a predictor is needs negative examples, and during robot development failures are rare.
Software engineering has long created such examples on purpose through fault injection, mutation testing and chaos engineering~\cite{faultinjection,mutationtesting,chaos}.
We inject faults during development for the same reason: to give \fail verdicts the labels they would otherwise rarely receive.

\section{Predictive Code Worlds}
\label{sec:method}

\ours has two parts.
The agent writes its skill on a \emph{code world} (\cref{sec:codeworld}), and the code world then makes \emph{world predictions} about the outcome of each subgoal (\cref{sec:worldpred}), which act only once they have been certified on injected faults (\cref{sec:calibrate}).
A code world whose world predictions have been certified in this way is what we call a predictive code world.
\Cref{fig:overview} follows our running example, putting a bowl into a drawer.

\subsection{Problem formulation}
\label{sec:setting}

\paragraph{Code-as-policy agents.}
A task is specified by an instruction.
The agent can reach the robot only through perception functions $\mathcal{P}$, such as segmentation, depth and oriented bounding boxes, and motion functions $\mathcal{U}$; it never reads simulator state.
The agent, a language model, writes a program $\pi$ that maps each observation $o$ to calls in $\mathcal{U}$.
It develops $\pi$ through trials on \emph{development} seeds, revises it from their execution records and keeps the version that passes most of them; disjoint \emph{held-out} seeds are never shown to it, and all results are reported on them.

\paragraph{What a program on perception leaves implicit.}
When $\pi$ is written directly against $\mathcal{P}$, every fact the task depends on is derived inside the control logic wherever it is needed, so the world the program acts in exists only implicitly, spread over its control flow.
And nothing in the program states what the world should look like after an action, so a failure that raises no error goes unnoticed.
\Cref{sec:codeworld} addresses the first gap, and \cref{sec:worldpred,sec:calibrate} the second.

\subsection{Code worlds}
\label{sec:codeworld}

A world model answers one question: \emph{after this action, what should the world look like?}
\ours lets the agent answer it for its own task, in code that runs on the robot's observations.

\paragraph{Definition.}
A code world $W=(\sigma, f, g)$ consists of three functions that the agent writes (\cref{fig:overview}a).
The state map $\sigma$ turns an observation into named facts, $s=\sigma(o)$, such as which objects are present, how they relate to the gripper and how far a drawer has been pulled; each fact carries a value, a unit, a coordinate frame and the perception evidence it was computed from.
The effect model $f$ predicts how the facts should change under a motion call $u\in\mathcal{U}$, $\hat{s}'=f(s,u)$; we call its outputs \emph{action-level predictions}, for instance that a grasped bowl should follow the gripper.
The goal map $g$ judges from $s$ whether the instruction is satisfied and how far the state is from it.
\paragraph{Writing the skill on a code world.}
The skill $\pi$ is written over $W$, and the code world serves it in four ways.
\emph{Facts.} The skill reads facts from $\sigma$ instead of calling $\mathcal{P}$: it asks how far the drawer is open, not which mask is the handle and where its points lie. Each quantity is derived once, with its unit and frame, instead of wherever the control logic needs it, so an error in a measurement is fixed in one place.
\emph{Expected effects.} After a motion call, the skill compares the new facts with $f$ and measures again when they disagree, so an execution slip is noticed at the call where it happens.
\emph{Rehearsal.} Before trials in the environment are spent, the agent runs candidate skills on $f$ and $g$ and discards those that do not reach the goal in the code world itself.
\emph{Graded progress.} The goal map tells development how far a trial got, rather than only whether it succeeded.
Together they change what the agent revises from: a failed trial is described in the code world's terms, which fact departed from its expected effect and how far away the goal was, rather than as a trace of perception outputs.

\paragraph{Why code, and why per task.}
A code world generates no images and learns no dynamics from data: what it holds, and what it predicts, are measurable facts about the scene and about the effects of the agent's own actions (the blue pictures in \cref{fig:teaser} are drawings of these facts).
Its facts are values that the skill can branch on, and its functions run on the robot's observations, so every one of them can be checked against what the robot then sees.
Because the agent writes it for one task, it holds exactly the quantities that task depends on, the opening of this drawer and the place of this bowl, and nothing else.
Whether skills written on a code world are better than skills the agent writes without one is what the ablation in \cref{sec:step1} tests.

\subsection{World predictions}
\label{sec:worldpred}

Action-level predictions are local: they say where the hand should be after a call, not whether the subgoal the call serves has been achieved.
A drawer can open exactly as predicted and be pushed back before the bowl arrives.
To catch such failures, the code world must predict outcomes at the level of \emph{subgoals}, the steps into which the agent divides its skill, such as opening the drawer, holding the bowl and placing it.

\paragraph{Definition.}
A world prediction $c=(g_c, m_c, \mathcal{S}_c, \mathcal{F}_c)$ has four parts (\cref{fig:overview}b):
the subgoal $g_c$;
code $m_c$ that measures, from the state, the quantity that tells success from failure;
the range $\mathcal{S}_c$ of that quantity expected after success;
and the range $\mathcal{F}_c$ expected after the most likely failure, disjoint from $\mathcal{S}_c$.
It is settled on a fresh observation $o'$ taken after the subgoal's last action:
\begin{equation}
v_c(o') =
\begin{cases}
\pass & \text{if } m_c(\sigma(o')) \in \mathcal{S}_c,\\
\fail & \text{if } m_c(\sigma(o')) \in \mathcal{F}_c,\\
\unknown & \text{otherwise,}
\end{cases}
\label{eq:settle}
\end{equation}
so that insufficient evidence, for instance an occluded part, yields \unknown rather than a forced verdict.
A world prediction thus states, before the robot looks, what the world should be like once its subgoal is done (\cref{fig:teaser}, right).
For example, \texttt{drawer\_open} measures how far the handle has moved along the pull direction from its closed position.
Two properties matter below.
Because a world prediction names the range of the most likely failure as well as that of success, a \fail verdict is a measurement of that failure rather than the absence of success.
And because it is measured on the code world's shared state, a completed subgoal can be checked again at any later point.

\paragraph{When should world predictions be written?}
A check written together with the skill enters its control flow before anyone knows whether the check is right, and a wrong verdict can then block a motion that the skill would otherwise have made.
We therefore add world predictions only to a skill that already succeeds on development trials, and let them change its behavior only through certified verdicts (\cref{sec:calibrate}).

\subsection{Certifying world predictions with injected faults}
\label{sec:calibrate}

A world prediction is written by the agent that wrote the skill, and it shares that agent's blind spots.
We therefore judge each prediction by outcomes, not by its author's confidence: before a prediction may act, we estimate how often each of its verdicts is right.
The difficulty is that clean development trials rarely fail exactly where a prediction is settled, so its \fail verdicts would receive almost no labels.
We therefore create such failures on purpose.

\paragraph{What the labels must cover.}
By \cref{eq:settle}, a verdict depends only on the state the code world measures, not on how that state came about: \texttt{drawer\_open} measures how far the drawer is open, and it cannot tell a drawer pushed back right after it was opened from one pushed back later while the bowl is carried.
Labels therefore need to cover failure \emph{states}, not failure \emph{causes}.
We inject a few faults into development trials that drive the world into the failure range of each prediction, such as dropping a grasped object or moving an operated part back right after it was operated.
The first verdict after a fault is labeled by the injected fault; otherwise, each prediction contributes only its last verdict in a trial, labeled from the outcome of the trial and, when the trial fails, from the later verdicts, which locate the failure.
Besides the injected faults, the outcome of a trial is the only ground truth available during development; the supplement gives the full rule.
A verdict certified on these labels can then also judge failures with other causes, as long as they end in the same states.

\paragraph{Certification rule.}
The \pass and \fail verdicts of a prediction $c$ are certified separately (\cref{fig:overview}c).
With $n_{c,d}$ labeled verdicts of kind $d\in\{\pass,\fail\}$, of which a fraction $\hat{p}_{c,d}$ is correct, verdict $d$ is \emph{certified} if
\begin{equation}
n_{c,d} \ge n_{\min}
\quad\text{and}\quad
\hat{p}_{c,d} \ge \gamma_d .
\label{eq:certify}
\end{equation}
The rule decides who may act and is not an estimate of probability: a verdict that has not shown its precision on labeled verdicts leaves the program exactly as it was.
It is deliberately simple. A verdict receives labels from only a few development trials, too few to support anything richer than a count, and the same rule with the same $n_{\min}$ and $\gamma_d$ applies to every prediction of every task: what differs between tasks is the predictions the agent writes and the labels the injected faults provide.
\pass is held to a stricter threshold than \fail, $\gamma_{\texttt{pass}} > \gamma_{\texttt{fail}}$: a wrong \fail costs a redo, whereas a wrong \pass carries an unachieved subgoal forward.
\Cref{sec:setup} gives the values of $n_{\min}$ and $\gamma_d$.

A verdict whose precision falls short of $\gamma_d$ also points the agent to a prediction, or a skill, that is wrong, and the agent revises it during development; \cref{fig:example} and \cref{sec:step2} trace one such revision.

\paragraph{Only certified verdicts act.}
\label{sec:act}
Once world predictions are added, the agent rewrites the skill into a loop over its subgoals that settles each subgoal's prediction on a fresh observation (\cref{fig:overview}d).
A certified \pass moves on, and a certified \fail triggers recovery.
Uncertified and \unknown verdicts leave the skill on its \emph{default path}, the one it takes with world predictions switched off, where every verdict is reported as \unknown, so a prediction that has not proved reliable cannot change what the program does.
A certified \fail says that a subgoal does not hold now, not when it was undone, so the program checks its completed subgoals again and rolls back to the earliest one that no longer holds.
Because every world prediction is measured on the shared state (\cref{sec:worldpred}), such a re-check costs one observation; the program also makes it before its last subgoal, which relies on all the earlier ones, and at the end.
A subgoal that a later step ends on purpose, such as holding the bowl once it is placed, then settles as \unknown and triggers nothing.
Like the skill, settling predictions and rolling back are plain steps of the program: at run time it queries no language model.
The whole development, including the addition of world predictions, can be repeated as \emph{self-improvement} (\cref{sec:setup}).

\section{Experiments}
\label{sec:experiments}

We ask three questions: how far \ours takes a small open-weight model on LIBERO-Pro and on two further benchmarks, what each of its two parts contributes (\cref{sec:step1}), and what world predictions change when episodes are disturbed (\cref{sec:step2}).

\subsection{Setup}
\label{sec:setup}

\paragraph{Benchmarks.}
We use LIBERO-Pro~\cite{liberopro}, which perturbs the LIBERO task suites~\cite{libero}, with the development and held-out seeds of \cref{sec:setting}, as in ASPIRE~\cite{aspire}.
Main results cover the 80 tasks on which ASPIRE reports results, and all success rates are on held-out seeds.
To test whether the gains carry over, we also run \ours on the two other benchmarks of ASPIRE, with its tasks and held-out seeds: seven Robosuite tasks~\cite{robosuite} with 100 seeds each, and two long-horizon household tasks of BEHAVIOR-1K~\cite{behavior1k}, navigating to and picking up a soda can and a radio, with 25 seeds each.

\paragraph{Compared systems.}
At the system level, we compare with ASPIRE: its published results with Claude Opus 4.6, Opus 5 and Opus 5.5, and its released code run with \qmax and with \flash, the model \ours uses.
Both systems can keep improving after a first working program, and our tables mark each row as with or without this \emph{self-improvement}.
ASPIRE self-improves with a skill library carried across tasks and several rounds of evolutionary search over programs.
Ours is similar in kind: an evolutionary search over programs with a memory shared by all tasks (details in the supplement).
We run ASPIRE with the default settings of its released code; ours uses five rounds.
Without self-improvement there is no such round: each task is developed once, from scratch.
At the component level, we remove the two parts of \ours one at a time: without either part, the agent writes its skill directly on the perception and motion APIs; with \emph{code world only}, it writes the skill on a code world; and the full \ours adds world predictions and the loop over subgoals that acts on their certified verdicts (\cref{sec:calibrate}).

\paragraph{Disturbances.}
On six LIBERO-Pro tasks from five suites, the evaluator disturbs held-out episodes; the program is not told which disturbance comes or when.
The disturbances fall into two groups; the supplement gives their triggers and parameters.
\emph{During an operation}, a grasped object is dropped, an object is displaced before it is grasped, or an operated part is moved back right after it was operated.
\emph{After confirmation}, an operated part is moved back silently after its subgoal has been confirmed, for instance a drawer pushed back while the bowl is carried to it.
Every variant meets the same disturbance on the same seed, and we compare variants with the exact McNemar test on the seeds where they differ.

\paragraph{Implementation.}
All our variants use \flash~\cite{qwen38} with the same limit on control steps per episode; held-out outcomes are never shown to the agent.
With \ours, the language model is used during development only: at evaluation, a program runs on perception and motion calls alone, world predictions included.
Perception uses SAM~3 segmentation~\cite{sam3}, depth, point clouds and oriented bounding boxes, and certification uses $n_{\min}=5$, $\gamma_{\texttt{pass}}=0.9$ and $\gamma_{\texttt{fail}}=0.8$ (\cref{eq:certify}) for every task.

\begin{figure*}[t]
\centering
\includegraphics[width=\linewidth]{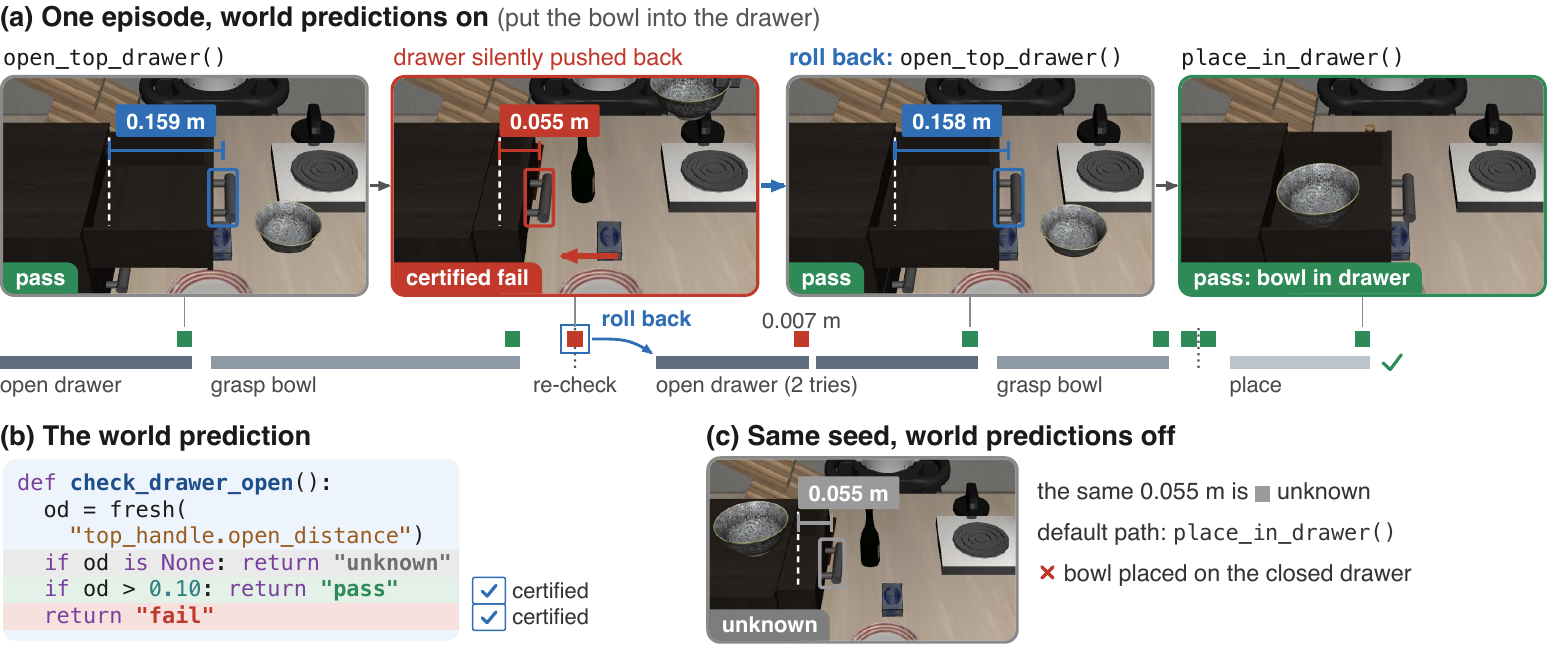}
\caption{\textbf{A certified world prediction at run time} (put the bowl into the drawer).
\textbf{(a)}~One held-out episode with world predictions on, and its loop as logged below the frames (squares are verdicts). After opening, \texttt{drawer\_open} measures 0.159\,m. The drawer is then silently pushed back while the bowl is held; the re-check before placing measures 0.055\,m, a certified \fail, and the program rolls back: it re-opens the drawer in two tries (the first measures 0.007\,m), grasps the bowl again and places it.
\textbf{(b)}~The world prediction: it passes above 0.10\,m, and both of its verdicts are certified, \fail on injected push-backs.
\textbf{(c)}~The same seed with world predictions off: the same 0.055\,m is \unknown, the program stays on its default path, and the bowl ends up on the closed drawer.}
\label{fig:example}
\end{figure*}

\begin{table}[t]
\centering
\small
\setlength{\tabcolsep}{4pt}
\caption{\textbf{Success on 80 LIBERO-Pro tasks} (held-out seeds, without our disturbances; same tasks for all rows).
ASPIRE~\cite{aspire} rows are ordered by model strength; \ours uses the weakest of these models.
Self-improving: see \cref{sec:setup}. Our rows use 50 held-out seeds per task.
ASPIRE's Opus rows are from~\cite{aspire} and average its eight suites (Opus 4.6: the two Long suites without evolutionary search).
Bold: ours. $^{\dagger}$Run with ASPIRE's released code.}
\label{tab:main}
\begin{tabular}{@{}llcc@{}}
\toprule
System & Model & Self-improving & Success (\%) \\
\midrule
ASPIRE & \flash & \xmark & 49.4$^{\dagger}$ \\
ASPIRE & \flash & \cmark & 63.8$^{\dagger}$ \\
ASPIRE & \qmax & \xmark & 58.3$^{\dagger}$ \\
ASPIRE & Claude Opus 4.6 & \cmark & 65.7 \\
ASPIRE & Claude Opus 5 & \cmark & 94.5 \\
ASPIRE & Claude Opus 5.5 & \cmark & 98.8 \\
\midrule
\ours (ours) & \flash & \xmark & \textbf{83.4} \\
\ours (ours) & \flash & \cmark & \textbf{89.4} \\
\bottomrule
\end{tabular}
\end{table}

\begin{table}[t]
\centering
\small
\setlength{\tabcolsep}{4pt}
\caption{\textbf{Ablation of the two parts} on the 80 tasks of \cref{tab:main}, with \flash and without self-improvement (50 held-out seeds per task). The last row is the full \ours; without a code world, the skill is written directly on the perception and motion APIs.}
\label{tab:ablation}
\begin{tabular}{@{}ccc@{}}
\toprule
Code world & World predictions & Success (\%) \\
\midrule
\xmark & \xmark & 49.4 \\
\cmark & \xmark & 78.2 \\
\cmark & \cmark & \textbf{83.4} \\
\bottomrule
\end{tabular}
\end{table}

\begin{table}[t]
\centering
\small
\setlength{\tabcolsep}{3pt}
\caption{\textbf{Success (\%) on Robosuite and BEHAVIOR-1K} (held-out seeds; the tasks and seeds of ASPIRE~\cite{aspire}): seven Robosuite tasks with 100 seeds each and two BEHAVIOR-1K tasks with 25 seeds each. ASPIRE numbers are from~\cite{aspire}. Bold: ours.}
\label{tab:other}
\setlength{\tabcolsep}{4pt}
\begin{tabular}{@{}lccc@{}}
\toprule
Model & Self-improving & Robosuite & BEHAVIOR-1K \\
\midrule
\multicolumn{4}{@{}l}{ASPIRE} \\
Claude Opus 4.6 & \cmark & 81.0 & 88.0 \\
Claude Opus 5 & \cmark & 98.9 & 96.0 \\
Claude Opus 5.5 & \cmark & 100.0 & 100.0 \\
\midrule
\multicolumn{4}{@{}l}{\ours (ours)} \\
\flash & \xmark & \textbf{88.3} & \textbf{90.0} \\
\flash & \cmark & \textbf{95.1} & \textbf{96.0} \\
\bottomrule
\end{tabular}
\end{table}

\subsection{Main results}
\label{sec:step1}

\paragraph{Comparison with ASPIRE.}
\Cref{tab:main} compares complete systems.
ASPIRE's success follows the strength of its model: without self-improvement, 49.4\% with \flash and 58.3\% with the larger \qmax; with it, 63.8\% with \flash, 65.7\% with Claude Opus 4.6, 94.5\% with Opus 5 and 98.8\% with Opus 5.5.
With \flash, the weakest of these models, \ours reaches 83.4\% without self-improvement: 34.0 points above ASPIRE with the same model, and 17.7 points above ASPIRE with Claude Opus 4.6, which self-improves.
Self-improvement raises it further to 89.4\%, 25.6 points above ASPIRE with the same model and its own self-improvement; the supplement gives the results per suite.
It is also cheaper: developing a task with \ours costs about \$3.9 on average at list prices, against about \$92, \$50 and \$15 for ASPIRE with Claude Opus 4.6, Opus 5 and Opus 5.5~\cite{aspire}.

\paragraph{What each part contributes.}
\Cref{tab:ablation} ablates the two parts of our method, without self-improvement; each does a different job.
The code world accounts for most of the gain: writing the skill on it raises success from 49.4\% to 78.2\%.
The two variants share the model, the tools and the way programs are developed, and differ only in what the skill is written on (\cref{sec:codeworld}), so this gain comes from the representation alone.
World predictions then add 5.2 points, to 83.4\%.
These episodes are undisturbed; under disturbances the gain is larger (\cref{sec:step2}).
The code world thus improves the skill that the model writes, and world predictions make the execution of that skill more robust.

\paragraph{What a failed trial tells the agent.}
The code world also changes what the agent learns from a failure (\cref{sec:codeworld}).
In one failed development trial of ``turn on the stove and put the moka pot on it'', the record returned to the agent is written in the code world's terms: the stove is on, the pot is 0.34\,m from the burner, progress stops at 0.21, and for each of the 423 steps it lists the predicted fact, the observed one and whether they match.
From this record the agent traces the failure to the pot's position, which is lost while the pot is carried, and revises how the pot is tracked in the hand and where it is grasped.
The next version succeeds on the same seed, with the pot 0.095\,m from the burner, and succeeds on 7 of the 15 development seeds instead of 3.

\paragraph{Beyond LIBERO-Pro.}
\Cref{tab:other} repeats the comparison with ASPIRE on Robosuite and BEHAVIOR-1K.
The same small model reaches 88.3\% and 90.0\% without self-improvement, and 95.1\% and 96.0\% with it.
Without self-improvement, this is 7.3 and 2.0 points above ASPIRE with Claude Opus 4.6; with it, 14.1 and 8.0 points, and level with ASPIRE with Claude Opus 5 on BEHAVIOR-1K.
Without self-improvement, world predictions add 1.6 and 4.0 points here (see the supplement).

\begin{figure}[t]
\centering
\includegraphics[width=\linewidth]{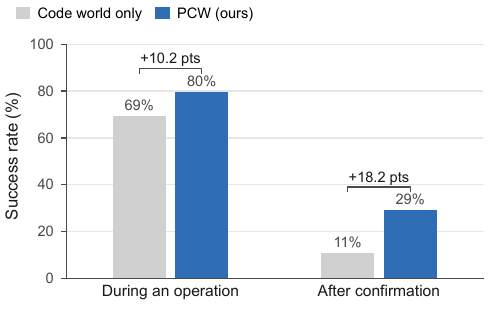}
\caption{\textbf{Robustness under disturbances on held-out seeds} (six tasks, the same disturbance on the same seed for both variants).
Bars show success rate, and brackets the gain of \ours in percentage points.}
\label{fig:main}
\end{figure}

\subsection{World predictions under disturbances}
\label{sec:step2}

\Cref{fig:main} compares code world only with \ours under the two groups of disturbances.
\emph{During an operation}, world predictions raise success from 69\% to 80\%, which removes a third of the failures.
\emph{After confirmation}, when an operated part is moved back after its subgoal has been confirmed, they nearly triple success, from 11\% to 29\%.
\ours also checks subgoals again and redoes them (\cref{sec:calibrate}), so we ask whether checking and retrying is enough, or whether the gain needs the predictions and their certification.

\paragraph{The recovery is made at run time, by certified verdicts.}
With world predictions switched off at run time, the same programs succeed in only 14\% of the episodes disturbed after confirmation, so the recovery is not a result of developing these programs further.
In most of the recovered episodes, a certified \fail rolls the program back to the undone subgoal.
Both gains of \cref{fig:main} and this difference are significant under the exact McNemar test ($p<10^{-7}$).

\paragraph{Without injected faults, there is almost nothing to roll back on.}
These \fail verdicts are certified only because of injected faults.
Clean development trials almost never fail where a prediction is settled: of the 42 world predictions in these programs, clean trials alone certify the \fail verdict of only one, and injected faults raise this to 13.
Without them, the programs would hold the same predictions and the same loop, and could roll back on almost none.

\paragraph{Certified verdicts hold on held-out seeds.}
Verdicts certified on development seeds stay reliable on held-out ones.
Under disturbances during an operation, every certified \fail settled right after a disturbance is correct, so the rollbacks that a disturbance triggers act on a correct signal, and no certified \fail fires where a subgoal is known to hold.
Certified \pass verdicts agree with the outcome of their episode 96.6\% of the time, against 70.6\% for uncertified ones: certification keeps unreliable verdicts from acting.

\paragraph{How a world prediction catches an undone subgoal.}
\Cref{fig:example} follows one world prediction, \texttt{drawer\_open}, starting in development.
An injected push-back leaves the drawer open by about 0.055\,m, and the first version of the prediction, which passes above 0.05\,m, judges it open.
Its \pass verdict is therefore not certified, and the agent raises the threshold to 0.10\,m, after which both verdicts are certified (\cref{fig:example}b).
This revision is what makes the run-time recovery possible: a drawer pushed back while the bowl is carried now measures as a certified \fail (\cref{fig:example}a), a signal the program does not have with world predictions switched off (\cref{fig:example}c).
Checking and retrying alone would not have been enough: under the first version of the prediction, the pushed-back drawer still counts as open.

\begin{figure}[t]
\centering
\includegraphics[width=\linewidth]{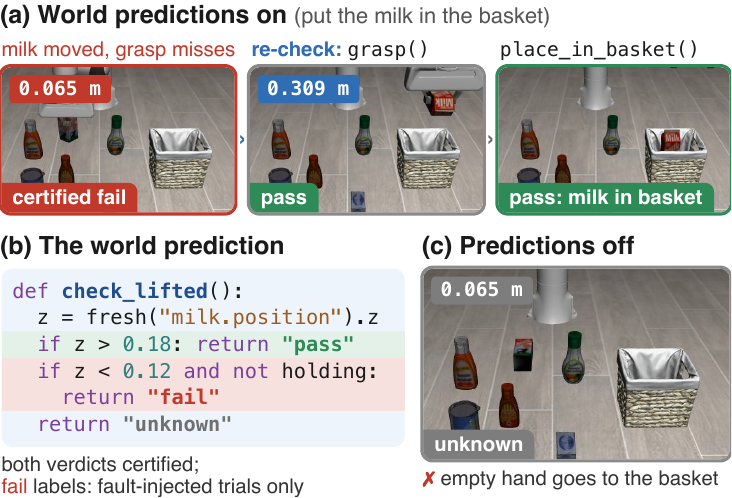}
\caption{\textbf{A second certified world prediction at run time} (put the milk in the basket).
\textbf{(a)}~The milk is moved before the grasp and the gripper closes on air. \texttt{lifted} measures the carton at 0.065\,m with an empty hand, a certified \fail; the program checks its earlier subgoals again, grasps the carton where it now is, and places it.
\textbf{(b)}~The world prediction.
\textbf{(c)}~Same seed, world predictions off: the same 0.065\,m is \unknown, the carton is never lifted, and an empty hand goes to the basket.}
\label{fig:milk}
\end{figure}

\paragraph{The same mechanism on another task.}
\Cref{fig:milk} shows a different task, a different disturbance and a different prediction: the milk is moved just before the grasp, \texttt{lifted} settles as a certified \fail, and the program looks again, lifts the carton and places it.
As with the drawer, this is possible only because of injected faults: clean development trials give this \fail verdict no label at all.

\section{Conclusion}
\label{sec:conclusion}

We presented \oursfull (\ours).
Before writing its policy, the agent builds from perception APIs a code world that holds the task's facts, the expected effect of each action and the progress toward the goal; it then adds predictions of each subgoal's outcome.
\ours thus turns two things that a small model does not do reliably by itself, working out the task's world and noticing when that world goes wrong, into code that the model writes, runs and revises.
The first makes its skills better.
The second lets its programs recover from disturbances, on one condition: a check that an agent writes for itself is worth acting on only after it has been tested against failures made on purpose.
The result is a lightweight agent: one small open-weight model is enough, developing a task costs a few dollars, and the program it writes runs without querying any language model.

\section*{LLM Usage Statement}
We used two LLM-based coding assistants, Claude Code (Anthropic) and Codex (OpenAI).
\emph{Writing:} they generated drafts of the text, including the related work section, and helped look up and check the references; the authors reviewed and edited all of it.
\emph{Code:} they generated most of the implementation and the experiment scripts, which the authors reviewed.
\emph{Figures and tables:} they generated their sources from the authors' designs and results.
\emph{Experiments:} they assisted in running and monitoring experiments and in analyzing results; the authors designed the experiments.
\emph{Ideas and claims:} they assisted in refining ideas and claims that the authors proposed.
No LLM was used to create or annotate data.
The agents studied in this paper are themselves driven by language models, which write the programs we evaluate: \flash for all variants of \ours, and \flash and \qmax for our runs of ASPIRE (\cref{sec:setup}).
The authors checked the results and the claims and are responsible for all content.

{
    \small
    \bibliographystyle{ieeenat_fullname}
    \bibliography{main}
}

\clearpage
\appendix
\setcounter{table}{0}
\setcounter{figure}{0}
\renewcommand{\thetable}{S\arabic{table}}
\renewcommand{\thefigure}{S\arabic{figure}}
\renewcommand{\theHtable}{S.\arabic{table}}
\renewcommand{\theHfigure}{S.\arabic{figure}}
\setlength{\csname @fptop\endcsname}{0pt}
\setlength{\csname @fpsep\endcsname}{16pt}
\section*{Appendix}
\noindent Sections, figures, tables, equations and references cited without a letter or an ``S'' are those of the main paper.

\section{Implementation details}
\label{sec:supp_impl}

\paragraph{Code base.}
\ours is built on the released code of ASPIRE~\cite{aspire}.
The simulation environments, the perception and motion APIs, the agent framework, the seed split, the limits and the success criterion are ASPIRE's, unchanged.
As in ASPIRE, \ours uses no privileged information: neither the agent nor its program reads the simulator's state, in development or at run time.
What \ours adds is stated where it appears: the code world and its run-time support (\cref{sec:supp_codeworld}), world predictions and their certification, and the blocks added to ASPIRE's prompt (\cref{sec:supp_tasks}).

\paragraph{Seeds.}
We keep ASPIRE's seeds and its protocol.
\Cref{tab:supp_seeds} gives the split of \cref{sec:setting} for each benchmark.
As in ASPIRE, the agent sees development seeds only, and the program that is frozen is selected on them; held-out seeds are kept from the agent, rejected by every simulation it starts, and used only to evaluate frozen programs.

\begin{table}[ht]
\centering
\small
\setlength{\tabcolsep}{4pt}
\caption{Seeds per task, as in ASPIRE~\cite{aspire}: seed ids, with their number in parentheses.}
\label{tab:supp_seeds}
\begin{tabular}{@{}lcc@{}}
\toprule
Benchmark & Development & Held-out \\
\midrule
LIBERO-Pro & 51--65 (15) & 1--50 (50) \\
Robosuite & 101--125 (25) & 1--100 (100) \\
BEHAVIOR-1K & 26--35 (10) & 1--25 (25) \\
\bottomrule
\end{tabular}
\end{table}

\paragraph{Model.}
All our variants and our runs of ASPIRE with \flash and with \qmax call the model in the same way: the same agent framework, Claude Code, which is the one ASPIRE uses, the same settings, and the server's default sampling parameters.

\paragraph{Limits.}
The limits on a development and on the control steps of an episode are those of ASPIRE's released code, the same for all our variants and for our runs of ASPIRE.
Retries and rollbacks count toward the same limit.

\paragraph{Self-improvement.}
This is the same mechanism as ASPIRE's (\cref{sec:setup}): an evolutionary search over programs in which each round develops every task again as in \cref{sec:setting} and starts from what earlier rounds left for all tasks.
This shared store is ASPIRE's skill library, in the same form; the paper calls it a memory.
Only what it holds differs, since each system fills it with summaries of the programs it wrote itself (\cref{sec:supp_tasks}, Prompts).
ASPIRE's self-improvement is run with the default settings of its released code.
Ours follows the same stopping rule and adds a cap, the five rounds of \cref{sec:setup}.

\paragraph{Labels.}
As in ASPIRE, the agent gets no privileged information: nothing gives it the simulator's state, in development or at run time.
This holds for everything \ours adds: the code world, the skill and the world predictions are written and run on what the APIs of \cref{tab:supp_api} return, and on nothing else.
Labels are assigned to verdicts after a development trial and serve to certify them (\cref{sec:calibrate}).
As in \cref{sec:calibrate}, the first verdict of the disturbed subgoal after an injected fault is labeled by the fault.
Otherwise, each prediction gets one label per trial, for its last verdict (\unknown gets none), inferred from the trial's outcome.
In a successful trial, a \pass is correct and a \fail is wrong.
In a failed trial, a \fail is correct unless a later prediction passes, a \pass of an intermediate subgoal is correct only if a later prediction fails, and a \pass of the final subgoal is wrong.
Trials with an injected fault are ordinary development trials and count toward the same development limit; the agent receives their record and, for each verdict, whether it is certified.
The thresholds of \cref{eq:certify} (\cref{sec:setup}) were set once, before any held-out episode was run.
How well certified verdicts hold on held-out seeds is reported in \cref{sec:step2}.

\paragraph{Run-time loop.}
By default, the loop over subgoals (\cref{sec:act}) tries each subgoal at most three times and rolls back at most twice per episode.
An \unknown or uncertified verdict changes nothing: after one more look, the program stays on its default path, where the skill's own confirmation decides whether to retry or to move on.
With world predictions switched off, every verdict is \unknown (\cref{sec:act}), so the program takes exactly this path.
As in ASPIRE, a frozen program is code on the perception and motion APIs of \cref{tab:supp_api} and nothing else.
The loop over subgoals is part of that code: it calls the same APIs and adds no sensing.
None of the interfaces in \cref{tab:supp_api} is a language model, so the program queries none at run time.

\section{Code world details}
\label{sec:supp_codeworld}

A code world is one Python module written on ASPIRE's perception and motion APIs; it adds no sensing: every fact in it is computed from what those APIs return.
\Cref{tab:supp_parts} maps the functions the agent writes to $W=(\sigma,f,g)$ of \cref{sec:codeworld}, and \cref{fig:supp_code} shows excerpts from the code world of the task of \cref{fig:example}, put the bowl into the drawer.

\begin{table}[ht]
\centering
\small
\setlength{\tabcolsep}{4pt}
\caption{The parts of a code world.}
\label{tab:supp_parts}
\begin{tabular}{@{}llp{0.47\linewidth}@{}}
\toprule
Part & Function & Role \\
\midrule
$\sigma$ & \texttt{observe}, \texttt{update} & write perception results as facts, with frame, unit and evidence \\
$f$ & \texttt{predict} & what the next observation should show after a motion call \\
$g$ & \texttt{done}, \texttt{progress} & whether the goal holds, and how far the state is from it \\
rehearsal & \texttt{simulate} & a return value for every API call from the effect model, without the simulator \\
\bottomrule
\end{tabular}
\end{table}

\begin{figure}[ht]
\scriptsize
\begin{verbatim}
# fact: how far the handle is from its closed
# position along the pull direction
state.set_fact("top_handle", "open_distance", od,
    frame="world", unit="m", evidence_ids=ids)

# effect model: after goto_pose the hand is at its
# target; a clamped handle follows the pull
def predict(call):
    if call["function"] == "goto_pose":
        preds = [dict(entity="hand", fact="position",
                      expected=hand_target,
                      tolerance=0.06)]
        if holding_handle and pull_along > 0.004:
            preds.append(dict(
                entity="top_handle", fact="position",
                expected=handle + step * pull_dir,
                tolerance=0.05))
        return {"predictions": preds}

# goal: both conditions need observed evidence
def done():
    return all_of(open_distance > 0.10,
                  bowl.in_drawer)

# the skill as a loop over subgoals
result = sl.run(world, [
    sl.State("drawer_open", act=act_open,
             observe=observe_open,
             confirm=confirm_open),
    sl.State("bowl_held", ...),
    sl.State("bowl_in_drawer", ...),
], max_tries=3, max_rollbacks=2)

def confirm_open():  # reads a fact of the world
    return known(od) and od["value"] > 0.10

# world prediction
CHECKS["drawer_open"] = {
  "success_looks_like": "handle >= 0.10 m from "
     "its closed baseline along the pull direction",
  "failure_looks_like": "back near the baseline or "
     "stalled half-open (pull missed, clamp "
     "slipped, or pushed back)"}

def check_drawer_open():
    # only a measurement after the last action
    od = fresh("top_handle.open_distance")
    if od is None:
        return "unknown"
    return "pass" if od > 0.10 else "fail"
\end{verbatim}
\caption{Excerpts from the code world, the skill and one world prediction for ``open the top layer of the drawer and put the bowl inside'' (shortened), from the final program of \cref{fig:example}. The motion call it predicts, \texttt{goto\_pose}, is one of ASPIRE's APIs (\cref{tab:supp_api}).}
\label{fig:supp_code}
\end{figure}

\paragraph{How the skill reads the code world.}
The skill's confirmations read facts, such as \texttt{top\_handle.open\_distance}, not masks or points.
\texttt{progress} grades a trial by what was observed of the bowl: seen, held, carried to the drawer, in the drawer.

\paragraph{Rehearsal.}
Every revised candidate is first run offline: the same skill code runs with each API call answered by \texttt{simulate}, and is also replayed against recorded development trials when there are any.
A candidate that raises an error does not reach the simulator.
Rehearsal never starts the simulator, so it adds no trial on any seed.

\paragraph{One world prediction and its certification.}
\Cref{tab:supp_gate} gives the certification record of \texttt{drawer\_open}.
After an injected push-back the drawer stays open by about 0.055\,m, which the first threshold of 0.05\,m judges open, so its \pass verdict is not certified.
With the threshold at 0.10\,m both verdicts are certified, and all eleven labels of its \fail verdict come from injected push-backs.

\begin{table}[ht]
\centering
\small
\setlength{\tabcolsep}{5pt}
\caption{Correct labels over all labels for the two verdicts of \texttt{drawer\_open}, before and after the agent revised its threshold.}
\label{tab:supp_gate}
\begin{tabular}{@{}lcc@{}}
\toprule
Threshold & \pass & \fail \\
\midrule
0.05\,m & 13/20, not certified & -- \\
0.10\,m & 38/41, certified & 11/11, certified \\
\bottomrule
\end{tabular}
\end{table}

\section{Tasks, APIs and prompts}
\label{sec:supp_tasks}
\paragraph{Tasks.}
The main results use the 80 LIBERO-Pro tasks of ASPIRE~\cite{aspire}: the Goal, Object, Spatial and Long suites, each under the position perturbation (\texttt{swap}) and the task perturbation (\texttt{task}), with ten tasks per suite.
\Cref{tab:supp_tasks} lists the six tasks of the disturbance study; all of them are among these 80.
They were drawn at random from the eight suites, before any disturbed episode was run.
On Robosuite and BEHAVIOR-1K we likewise use ASPIRE's tasks and held-out seeds (\cref{sec:setup}).

\begin{table}[ht]
\centering
\small
\caption{The six LIBERO-Pro tasks of the disturbance study, from five suites.}
\label{tab:supp_tasks}
\begin{tabular}{@{}lp{0.68\linewidth}@{}}
\toprule
Suite & Instruction \\
\midrule
\texttt{goal\_swap} & Open the top layer of the drawer and put the bowl inside \\
\texttt{goal\_task} & Open the top layer of the drawer and put the cream cheese inside \\
\texttt{10\_swap} & put the black bowl in the bottom drawer of the cabinet and close it \\
\texttt{10\_swap} & turn on the stove and put the moka pot on it \\
\texttt{object\_task} & Pick the milk and place it in the basket \\
\texttt{spatial\_task} & Pick the akita black bowl next to the cookie box and place it on the plate \\
\bottomrule
\end{tabular}
\end{table}

\paragraph{APIs.}
The observation, perception, grasping and motion interfaces are those of ASPIRE, unchanged, on all three benchmarks; \cref{tab:supp_api} lists them for LIBERO-Pro.
As in ASPIRE, none of them reads the simulator's internal state.
\ours adds only the last row, the run-time support of the code world: it stores facts and settles predictions computed from what the other interfaces return, and adds no perception or motion of its own.

\begin{table}[ht]
\centering
\small
\setlength{\tabcolsep}{4pt}
\caption{Interfaces a program may call on LIBERO-Pro: the perception and motion APIs of ASPIRE~\cite{aspire}, unchanged, and the run-time support of the code world that \ours adds. None reads the simulator's state.}
\label{tab:supp_api}
\begin{tabular}{@{}lp{0.74\linewidth}@{}}
\toprule
Kind & Interface \\
\midrule
\multicolumn{2}{@{}l}{\emph{As in ASPIRE}} \\
Observation & RGB, depth, camera parameters, arm state \\
Perception & SAM~3 segmentation from a text or point prompt; mask or pixel to world points; oriented box of a point cloud \\
Grasping & grasp candidates; choice of a top-down grasp \\
Motion & inverse kinematics; move to a pose or to joint values; open and close the gripper; go home \\
\midrule
\multicolumn{2}{@{}l}{\emph{Added by \ours}} \\
Code world & set and get facts; declare and settle world predictions; run the loop over subgoals \\
\bottomrule
\end{tabular}
\end{table}

\paragraph{Prompts.}
We start from the prompt of ASPIRE's released code and keep it as it is.
\ours adds to it, one block per part.
\emph{Code world}, added before the skill is written. The agent first defines the functions of \cref{tab:supp_parts}; facts must cite real observations, and parameters the development evidence they come from; every revised candidate is rehearsed offline first.
\emph{World predictions}, added once the skill succeeds on development trials. The agent writes for each subgoal what success looks like, what its typical failure is and which quantity is measured, and wraps the skill into the loop over subgoals of \cref{sec:act}.
\emph{Self-improvement}, as in ASPIRE. Each round uses the prompt with the code-world block again, with the shared library added: summaries of what earlier rounds wrote for all tasks (task analyses, findings and final programs), from development seeds only.
The agent is asked to treat them as evidence, not as correct, and may not reuse a program it has not tested again in the current round.
The blocks are the same text for every task.
Without a code world, the first row of \cref{tab:ablation}, the prompt is ASPIRE's alone: this row is the run of ASPIRE with \flash without self-improvement in \cref{tab:main}.
Our runs of ASPIRE use its prompt as released.

\section{Disturbances}
\label{sec:supp_dist}

\Cref{tab:supp_dist} gives the triggers and effects of the disturbances of \cref{sec:setup}.
The kinds are common failures of manipulation, fixed in advance, with the same rules for all tasks.
Faults in development (\cref{sec:calibrate}) are such failures on development seeds, where they label verdicts; the study asks whether the verdicts certified on them catch such failures on held-out seeds.
It keeps the held-out seeds, the APIs, the model settings and the limit on control steps of the main results on its six tasks; only the disturbance is added.
Like the ablation of \cref{tab:ablation}, it uses programs developed without self-improvement.
An episode is disturbed at most once.
The kind of disturbance and its parameters are assigned per seed and are the same for all compared variants; the disturbance is applied when its trigger in \cref{tab:supp_dist} is met.
Nothing is disturbed after the simulator has reported success, and the program is never told: it can notice a disturbance only through the perception APIs.
\emph{During an operation}, each held-out seed is assigned one kind that applies to the task: a drop or a displacement when there is a movable object, a push-back when there is an articulated part.
\emph{After confirmation}, every held-out seed gets the same kind.
Faults during development are injected when world predictions are added, on half of the development seeds, and never on held-out seeds.

\begin{table}[ht]
\centering
\small
\setlength{\tabcolsep}{3pt}
\caption{Disturbances on held-out episodes.}
\label{tab:supp_dist}
\begin{tabular}{@{}lp{0.37\linewidth}p{0.37\linewidth}@{}}
\toprule
Kind & Trigger & Effect \\
\midrule
\multicolumn{3}{@{}l}{\emph{During an operation}} \\
Drop & the grasped object is first lifted by 3\,cm & the gripper is forced open and the object falls back \\
Displace & the fingers first come within 6\,cm of an object not yet grasped & the object moves 3--5\,cm in a random horizontal direction \\
Push-back & an articulated part first passes 60\% of its range and the gripper is 5\,cm away & the part is pushed back to 40\% \\
\midrule
\multicolumn{3}{@{}l}{\emph{After confirmation}} \\
Undo & a subgoal has held for 25 steps, the program works on the next one, and the gripper is 10\,cm away & a placed object is moved out of its goal region, or an articulated part is pushed back to 40\% \\
\bottomrule
\end{tabular}
\end{table}

\section{Results per suite and per task}
\label{sec:supp_breakdown}
\Cref{tab:supp_suites} breaks the LIBERO-Pro results of the main table down by suite, and \cref{tab:supp_robosuite,tab:supp_behavior} the Robosuite and BEHAVIOR-1K results by task.
All columns are on ASPIRE's tasks and held-out seeds, with the same APIs and limits.
The ASPIRE columns with \flash are our runs of its released code with its default settings and the model settings of \ours; the Opus columns are the numbers ASPIRE reports.
With the same model and without self-improvement, writing the skill on a code world is above ASPIRE on every LIBERO-Pro suite, and world predictions add to it on every suite.
With self-improvement, \ours with \flash is above ASPIRE with the same model and its own self-improvement on all eight LIBERO-Pro suites, and above ASPIRE with Claude Opus 4.6 on all eight, at or above it on all seven Robosuite tasks, and above it on both BEHAVIOR-1K tasks.
Even without self-improvement, \ours is above ASPIRE with the same model and its self-improvement on all eight LIBERO-Pro suites.
The largest differences are on the Spatial suites of LIBERO-Pro and on nut assembly in Robosuite.
Without self-improvement, world predictions add 1.6 points to code world only on Robosuite and 4 points on BEHAVIOR-1K.

\paragraph{Cost.}
The development cost of \cref{sec:step1}, about \$3.9 per task, covers all of \ours: the code world and the skill, the world predictions, and self-improvement.
With the same model, limits and list prices, developing a task with ASPIRE and its self-improvement costs about \$4.1, as its agent takes more images into its context.

\begin{table}[ht]
\centering
\small
\setlength{\tabcolsep}{4pt}
\caption{\textbf{Success (\%) per LIBERO-Pro suite} (ASPIRE's tasks and held-out seeds; ten tasks per suite, 50 seeds per task). The Opus 4.6 numbers are from~\cite{aspire}; \texttt{swap} is its position perturbation and \texttt{task} its task perturbation. Bold: best in each row. $^{\dagger}$Run with ASPIRE's released code. $^{\ddagger}$Without evolutionary search: ASPIRE reports Opus 4.6 on the two Long suites only without it (\cite{aspire}, Fig.~9 and Tab.~24).}
\label{tab:supp_suites}
\resizebox{\linewidth}{!}{%
\begin{tabular}{@{}lcccccc@{}}
\toprule
 & \multicolumn{3}{c}{ASPIRE} & \multicolumn{3}{c}{Ours} \\
\cmidrule(lr){2-4}\cmidrule(l){5-7}
Model & \multicolumn{2}{c}{Qwen3.8-Flash$^{\dagger}$} & Opus 4.6 & \multicolumn{3}{c}{\flash} \\
\cmidrule(lr){2-3}\cmidrule(l){5-7}
 & & & & Code world & & \\
 & & & & only & \ours & \ours \\
Self-improving & \xmark & \cmark & \cmark & \xmark & \xmark & \cmark \\
\midrule
\texttt{object\_swap} & 88.0 & 96.4 & 98.0 & 89.4 & 97.8 & \textbf{99.2} \\
\texttt{object\_task} & 83.2 & 91.2 & 95.0 & 96.8 & \textbf{97.0} & \textbf{97.0} \\
\texttt{goal\_swap} & 64.0 & 83.0 & 81.0 & 85.4 & 85.8 & \textbf{98.6} \\
\texttt{goal\_task} & 39.0 & 44.0 & 45.0 & 70.4 & 74.4 & \textbf{78.2} \\
\texttt{spatial\_swap} & 28.0 & 47.0 & 51.0 & 90.8 & 92.4 & \textbf{97.8} \\
\texttt{spatial\_task} & 41.0 & 62.2 & 60.0 & 92.6 & 96.8 & \textbf{97.2} \\
\texttt{10\_swap} & 21.8 & 37.4 & 43.2$^{\ddagger}$ & 47.0 & 56.8 & \textbf{74.8} \\
\texttt{10\_task} & 30.4 & 48.8 & 52.0$^{\ddagger}$ & 53.4 & 66.2 & \textbf{72.0} \\
\midrule
All 80 tasks & 49.4 & 63.8 & 65.7 & 78.2 & 83.4 & \textbf{89.4} \\
\bottomrule
\end{tabular}%
}
\end{table}

\begin{table}[ht]
\centering
\small
\setlength{\tabcolsep}{4pt}
\caption{\textbf{Success (\%) per Robosuite task} (ASPIRE's tasks and held-out seeds; 100 seeds per task). ASPIRE numbers are from~\cite{aspire}. Bold: best in each row.}
\label{tab:supp_robosuite}
\begin{tabular}{@{}lcccc@{}}
\toprule
 & \multicolumn{3}{c}{Ours, \flash} & ASPIRE \\
\cmidrule(lr){2-4}
 & Code world & & & \\
 & only & \ours & \ours & Opus 4.6 \\
Self-improving & \xmark & \xmark & \cmark & \cmark \\
\midrule
\texttt{cube\_lift} & 94 & \textbf{100} & \textbf{100} & 97 \\
\texttt{cube\_stack} & 99 & \textbf{100} & \textbf{100} & 99 \\
\texttt{cube\_restack} & \textbf{100} & \textbf{100} & \textbf{100} & \textbf{100} \\
\texttt{spill\_wipe} & 99 & \textbf{100} & \textbf{100} & 99 \\
\texttt{nut\_assembly} & 64 & 64 & \textbf{74} & 9 \\
\texttt{two\_arm\_lift} & 67 & 67 & \textbf{100} & 71 \\
\texttt{two\_arm\_handover} & 84 & 87 & \textbf{92} & \textbf{92} \\
\midrule
Mean & 86.7 & 88.3 & \textbf{95.1} & 81.0 \\
\bottomrule
\end{tabular}
\end{table}

\begin{table}[ht]
\centering
\small
\setlength{\tabcolsep}{4pt}
\caption{\textbf{Success (\%) per BEHAVIOR-1K task} (ASPIRE's tasks and held-out seeds; 25 seeds per task). ASPIRE numbers are from~\cite{aspire}. Bold: best in each row.}
\label{tab:supp_behavior}
\begin{tabular}{@{}lcccc@{}}
\toprule
 & \multicolumn{3}{c}{Ours, \flash} & ASPIRE \\
\cmidrule(lr){2-4}
 & Code world & & & \\
 & only & \ours & \ours & Opus 4.6 \\
Self-improving & \xmark & \xmark & \cmark & \cmark \\
\midrule
Soda can & 88 & 92 & \textbf{96} & 88 \\
Radio & 84 & 88 & \textbf{96} & 88 \\
\midrule
Mean & 86.0 & 90.0 & \textbf{96.0} & 88.0 \\
\bottomrule
\end{tabular}
\end{table}

\end{document}